\documentclass[10pt,journal]{IEEEtran}
\usepackage{ifpdf}
\usepackage{amsthm,amsmath,amssymb,array}
\usepackage[caption=false,font=footnotesize]{subfig}
\usepackage{arydshln}
\usepackage{mathrsfs}

\ifCLASSOPTIONcompsoc
  \usepackage[nocompress]{cite}
\else
  \usepackage{cite}
\fi
\usepackage{multirow}
\ifCLASSINFOpdf
  \usepackage[pdftex]{graphicx}
\else
  \usepackage[dvips]{graphicx}
\fi
\usepackage{algorithmic}

\usepackage{times}
\usepackage{epsfig}
\usepackage{graphicx}
\usepackage{amsmath}
\usepackage{amssymb}
\usepackage[ruled]{algorithm2e}
\usepackage{amsmath}
\usepackage{makecell}
\usepackage{multirow}
\usepackage{color}
\usepackage[pagebackref=true,breaklinks=true,letterpaper=true,colorlinks,bookmarks=false]{hyperref}

\ifCLASSINFOpdf
\else
\fi
\begin{document}
%
% paper title
% Titles are generally capitalized except for words such as a, an, and, as,
% at, but, by, for, in, nor, of, on, or, the, to and up, which are usually
% not capitalized unless they are the first or last word of the title.
% Linebreaks \\ can be used within to get better formatting as desired.
% Do not put math or special symbols in the title.
\title{Conditional Hyper-Network for Blind Super-Resolution with Multiple Degradations}

% author names and affiliations
% transmag papers use the long conference author name format.

% \author{\IEEEauthorblockN{Guanghao Yin\IEEEauthorrefmark{1}$^{*}$\thanks{Equal contribution},
% Wei Wang\IEEEauthorrefmark{2}$^{*}$,
% Zehuan Yuan\IEEEauthorrefmark{2},
% Dongdong Yu\IEEEauthorrefmark{2},
% Shouqian Sun\IEEEauthorrefmark{1} and
% Changhu Wang\IEEEauthorrefmark{2},~\IEEEmembership{Fellow,~IEEE}}
% \IEEEauthorblockA{\IEEEauthorrefmark{1}Zhejiang University, Hangzhou, China}
% \IEEEauthorblockA{\IEEEauthorrefmark{2}Bytedance AI Lab, Shanghai, China}
% % \IEEEauthorblockA{\IEEEauthorrefmark{3}Starfleet Academy, San Francisco, CA 96678 USA}
% % \IEEEauthorblockA{\IEEEauthorrefmark{4}Tyrell Inc., 123 Replicant Street, Los Angeles, CA 90210 USA}% <-this % stops an unwanted space
% % \thanks{Manuscript received December 1, 2012; revised August 26, 2015.
% % Corresponding author: M. Shell (email: http://www.michaelshell.org/contact.html).}
% }
\author{Guanghao~Yin$^{*}$, Wei~Wang$^{*}$, Zehuan~Yuan, Wei~Ji, Dongdong~Yu, Shouqian~Sun, Tat-Seng~Chua, Changhu~Wang
% , ~\IEEEmembership{Senior Member, IEEE}

\IEEEcompsocitemizethanks{\IEEEcompsocthanksitem $^{*}$ Equal contribution.

Guanghao Yin, Shouqian Sun are with the Key Laboratory of Design Intelligence and Digital Creativity of Zhejiang Province, Zhejiang University, Hangzhou, China. E-mail: \{ygh\_zju, ssq\}@zju.edu.cn.

Wei Ji, Tat-seng Chua are with School of Computing, National University of Singapore, Singapore. E-mail: \{jiwei,dcscts\}@nus.edu.sg.

Wei Wang, Zehuan Yuan, Dongdong Yu, Changhu Wang are with the Bytedance AI Lab, Beijing, China.  E-mail: \{wangwei.frank, yuanzehuan, yudongdong, wangchanghu\}@bytedance.com.
}% <-this % stops an unwanted space
% \thanks{Manuscript received May 20, 2020; revised June 30, 2020.}
}

% The paper headers
\markboth{Journal of \LaTeX\ Class Files,~Vol.~14, No.~8, August~2015}%
{Shell \MakeLowercase{\textit{et al.}}: Bare Demo of IEEEtran.cls for IEEE Transactions on Image Processing}
% The only time the second header will appear is for the odd numbered pages
% after the title page when using the twoside option.
%
% *** Note that you probably will NOT want to include the author's ***
% *** name in the headers of peer review papers.                   ***
% You can use \ifCLASSOPTIONpeerreview for conditional compilation here if
% you desire.

% If you want to put a publisher's ID mark on the page you can do it like
% this:
%\IEEEpubid{0000--0000/00\$00.00~\copyright~2015 IEEE}
% Remember, if you use this you must call \IEEEpubidadjcol in the second
% column for its text to clear the IEEEpubid mark.

% use for special paper notices
%\IEEEspecialpapernotice{(Invited Paper)}

% for Transactions on Magnetics papers, we must declare the abstract and
% index terms PRIOR to the title within the \IEEEtitleabstractindextext
% IEEEtran command as these need to go into the title area created by
% \maketitle.
% As a general rule, do not put math, special symbols or citations
% in the abstract or keywords.
\IEEEtitleabstractindextext{%
\begin{abstract}
  Although the single-image super-resolution (SISR) methods have achieved great success on the single degradation, they still suffer performance drop with multiple degrading effects in real scenarios. Recently, some blind and non-blind models for multiple degradations have been explored. However, these methods usually degrade significantly for distribution shifts between the training and test data. Towards this end, we propose a novel conditional hyper-network framework for super-resolution with multiple degradations (named \textbf{CMDSR}), which helps the SR framework learn how to adapt to changes in the degradation distribution of input. We extract degradation prior at the task-level with the proposed ConditionNet, which will be used to adapt the parameters of the basic SR network (BaseNet). Specifically, the ConditionNet of our framework first learns the degradation prior from a support set, which is composed of a series of degraded image patches from the same task. Then the adaptive BaseNet rapidly shifts its parameters according to the conditional features. Moreover, in order to better extract degradation prior, we propose a task contrastive loss to shorten the inner-task distance and enlarge the cross-task distance between task-level features. Without predefining degradation maps, our blind framework can conduct one single parameter update to yield considerable improvement in SR results. Extensive experiments demonstrate the effectiveness of CMDSR over various blind, and even several non-blind methods. The flexible BaseNet structure also reveals that CMDSR can be a general framework for a large series of SISR models. Our code is available at \url{https://github.com/guanghaoyin/CMDSR}.
\end{abstract}

% Note that keywords are not normally used for peerreview papers.
}

% make the title area
\maketitle

% To allow for easy dual compilation without having to reenter the
% abstract/keywords data, the \IEEEtitleabstractindextext text will
% not be used in maketitle, but will appear (i.e., to be "transported")
% here as \IEEEdisplaynontitleabstractindextext when the compsoc
% or transmag modes are not selected <OR> if conference mode is selected
% - because all conference papers position the abstract like regular
% papers do.
\IEEEdisplaynontitleabstractindextext
% \IEEEdisplaynontitleabstractindextext has no effect when using
% compsoc or transmag under a non-conference mode.

% For peer review papers, you can put extra information on the cover
% page as needed:
% \ifCLASSOPTIONpeerreview
% \begin{center} \bfseries EDICS Category: 3-BBND \end{center}
% \fi
%
% For peerreview papers, this IEEEtran command inserts a page break and
% creates the second title. It will be ignored for other modes.
\IEEEpeerreviewmaketitle

\section{Introduction}
\label{Introduction}

Single image super-resolution (SISR) has posed a long-standing challenge in low-level vision with numerous important applications. It is an ill-posed problem that aims to restore a High-Resolution (HR) image by adding the missing high-frequency information from a Low-Resolution (LR) image. Since the pioneering method by SRCNN~\cite{dong2015image}, deep learning approaches~\cite{kim2016accurate,kim2016deeply,lim2017enhanced,zhang2018image, ledig2017photo, wang2018esrgan, tian2020coarse} have exhibited impressive performance. However, most existing methods focus on a fixed degradation, $i.e$, bicubic down-sampling or single Gaussian blurring. Such fixed settings really limit their generalization ability. In addition to down-sampling, unknown blurring and noise may also be introduced during the acquisition of LR images. When the data distributions at test time mismatch the training distributions (referred to as distribution shift~\cite{quionero2009dataset,lazer2014parable}), these learning-based models will suffer severe performance drop~\cite{yang2014single}.

\begin{figure}
  \centering
  \includegraphics[width=8cm]{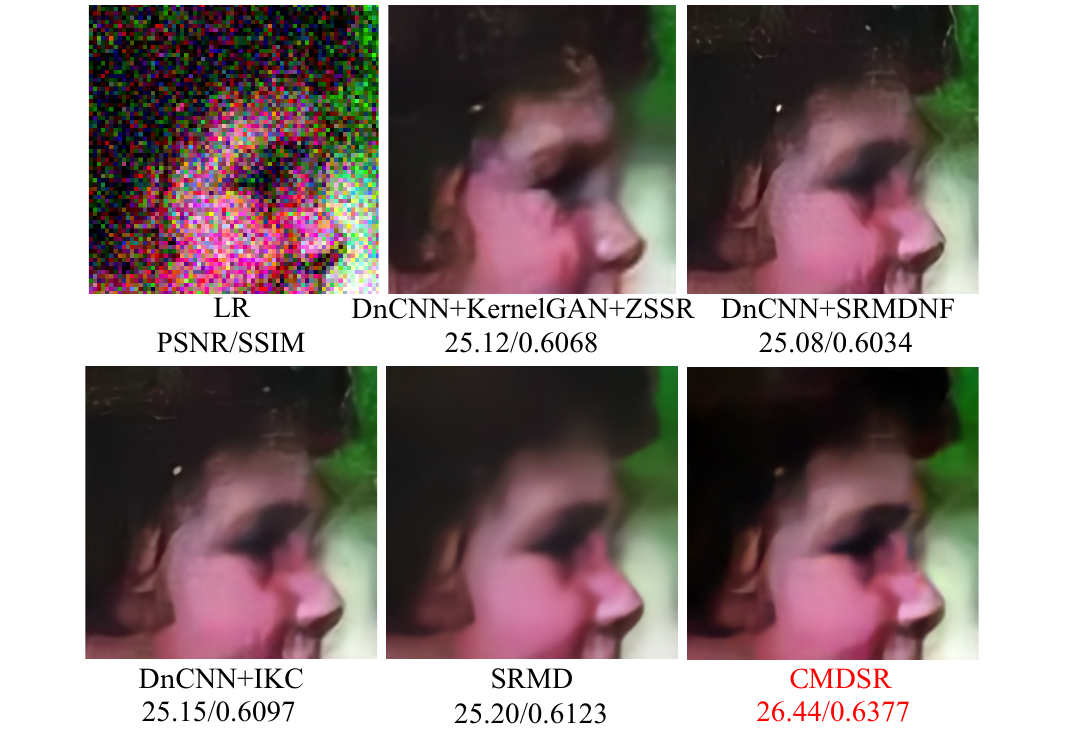}
  \caption{SR results ($\times 4$) with anisotropic Gaussian blur and AWGN (\textit{Severe} degradation). Without anisotropic blur kernel for training, our CMDSR outperforms all the blind cascaded schemes, even better than non-blind method, SRMD~\cite{zhang2018learning}, which was trained with anisotropic Gaussian blur.}
  \label{visual_anti_50}
\end{figure}

In recent years, several non-blind and blind super-resolution methods for multiple degradations have been proposed. The non-blind methods~\cite{riegler2015conditioned,zhang2018learning,zhang2019deep,xu2020unified} usually take the ground truth (GT) degradation maps as an additional input to establish the LR-HR mapping. Although the non-blind models have achieved satisfactory performance with the guidance of predefined information, the problem with unknown realistic degradation largely limits their usage in real-world applications. Besides, the blind methods~\cite{michaeli2013nonparametric,shocher2018zero,gu2019blind, bell2019blind} only consider the blur and down-sampling in the degradation mode. Then, the cascaded schemes with blind denoising, blur estimation and SR methods are organized to restore the multi-degraded LR image~\cite{zhang2018learning,liu2020learning,maeda2020unpaired}. However, each stage has a negative impact on each other, ($i.e$, the denoiser will  make the LR image more blurred and lead to kernel mismatch, increasing the difficulty of deblurring, as seen in Fig.~\ref{visual_anti_50_2}). Recently, there are some new attempts for blind SR. Several CycleGAN~\cite{zhu2017unpaired} based methods~\cite{bulat2018learn, yuan2018unsupervised, maeda2020unpaired,liu2020unsupervised} learn from unpaired LR-HR images, but they are more difficult to train. ZSSR~\cite{shocher2018zero} explores the zero-shot solution for the first time, where the CNN learns the mapping from the LR image and its downscaled versions (self-supervision). But it requires thousands of self-training iterations for each LR image. Recently, two optimization based meta-learning strategies, MZSR~\cite{soh2020meta} and MLSR~\cite{park2020fast}, have been proposed to accelerate the self-training steps from 1000 to 10. But they have limitations when dealing with large scale factor because the self-downsampled image cannot provide enough information. Moreover, all those blind and non-blind methods cannot handle the degradation shift problem. When degradation distribution shifts between the training and test data, those SOTAs usually degrade significantly.

To address the problems above, our work tries to explore the following two challenges at the same time: (1) Can we propose a blind framework to effectively handle multiple degradations, especially when the accurate degradation estimation is very difficult? (2) Is it possible to overcome the distribution shift with an adaptive model, which can learn how to adapt its parameters to the unknown degraded LR images?

\begin{figure}[t]
  \includegraphics[width=8.5cm]{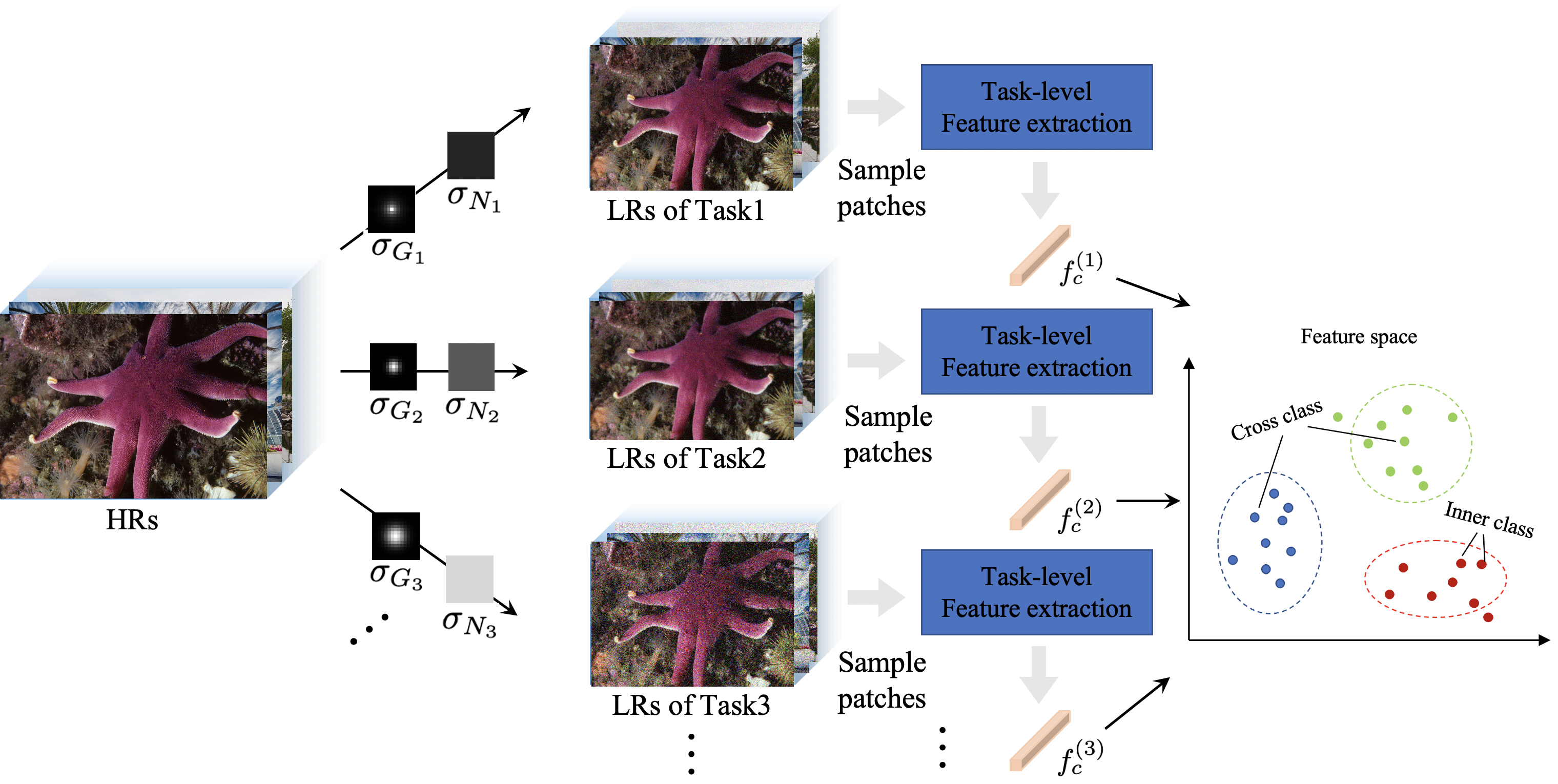}
  \centering
  \caption{Conditional feature extraction at task-level. There is a prior knowledge that degraded LR images from the same task have common degradation patterns ($i.e$, blur kernel width $\sigma_{G}$, noise level $\sigma_{N}$), which are different between different tasks. Therefore, we attempt to learn degradation prior at task-level and use the extracted conditional features to adapt the SR model to a new task.}
  \label{Task-level}
  % \vspace{-0.5cm}
\end{figure}

In this paper, we propose a conditional hyper-network for blind SR with multiple degradations (CMDSR) to largely overcome the aforementioned two challenges. For the first challenge, there is a prior knowledge which inspires us to handle it at task-level. As shown in Fig.~\ref{Task-level}, the LR images with different degradations obey different distributions. Although the accurate estimation of degradation is hard, images from the same task may contain similar implicit features to describe their common degradation patterns. Therefore, we group these LR images into different tasks and extract the degradation priors at task-level to describe the degradation patterns. Throughout this article, we define a group of LR patches with the same degradation pattern as a \textbf{support set}. For the second challenge, we use the distribution information extracted from the support set to make SR network adaptively adjust its parameters according to the distribution changes, such that our framework can handle distribution shifts.

\begin{figure*}[ht]
  \centering
  \includegraphics[width=16cm]{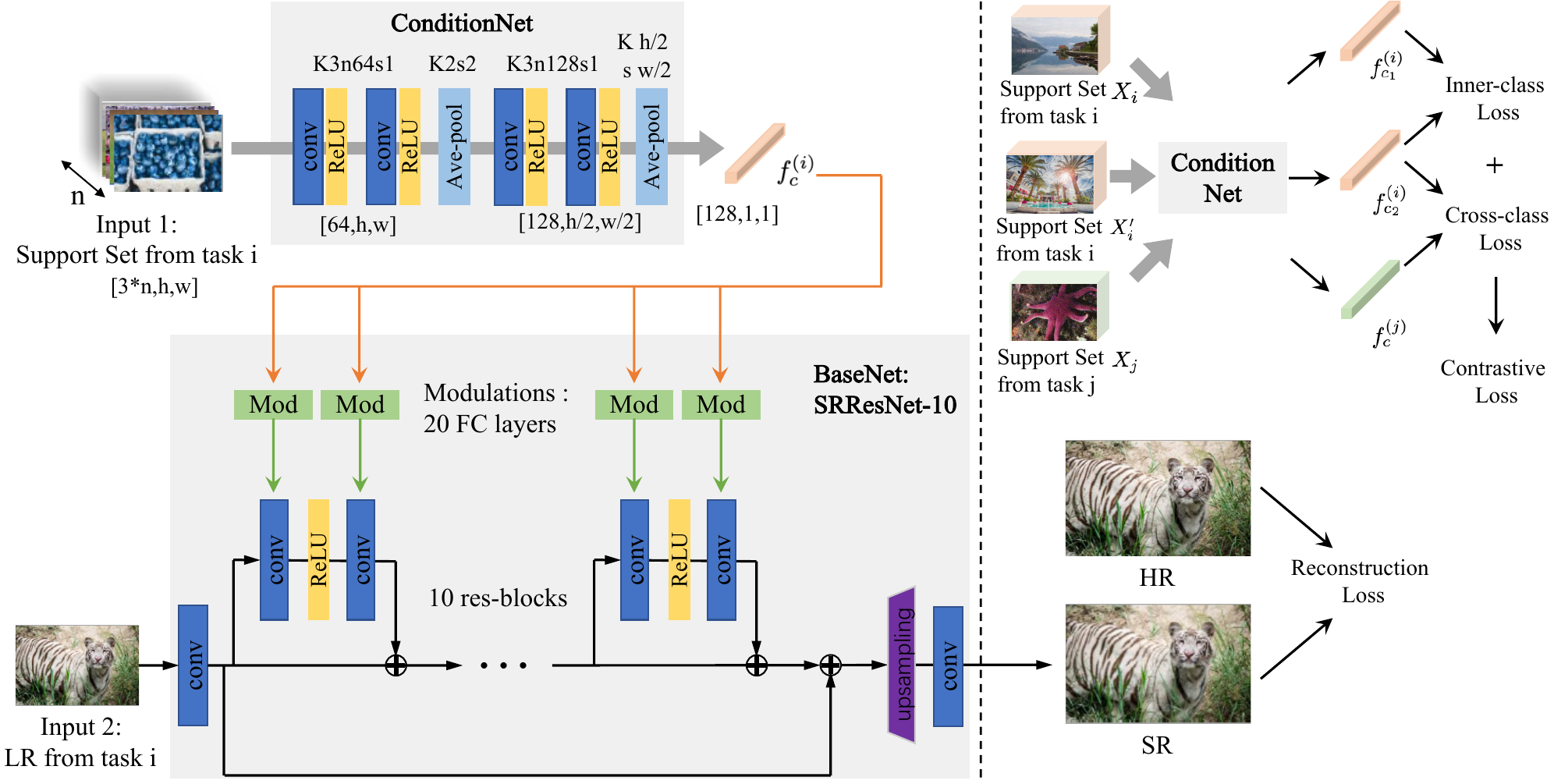}
  \caption{Overall scheme of our proposed CMDSR. Left: Network structures of ConditionNet and BaseNet. ConditionNet extracts conditional feature $f_c^{(i)}$ from input support set. BaseNet adapts its parameters to task $i$ according to the modulated features. Right: Loss functions optimization. BaseNet and ConditionNet are alternately trained by different loss configurations. It should be noted that although we simply use SRResNet-10 as BaseNet, CMDSR has no strict restrictions on the BaseNet structure and can be conveniently extended to other SR models without changing their structures.
  }
  \label{CMDSR}
  % \vspace{-0.5cm}
\end{figure*}

Specifically, our CMDSR consists of two parts: the BaseNet and ConditionNet. As shown in Fig.~\ref{CMDSR}, the shallow ConditionNet learns the feature representations of different tasks. Then, BaseNet multiplies its convolution weights with modulated conditional features in channel-wise. Finally, the adapted BaseNet restores the LR image. Inspired by recent contrastive learning~\cite{hadsell2006dimensionality,he2020momentum}, we propose a task contrastive loss to decrease the distances of the conditional features from the same task and increase the distances of the conditional features from different tasks. Algorithm \ref{CMDSR Training} presents the training stage, where BaseNet and ConditionNet are alternately optimized with different steps and loss functions. Algorithm~\ref{CMDSR Test} presents the test stage, where the extracted degradation prior from ConditionNet adapts BaseNet to handle distribution shift.

Since the shallow ConditionNet only uses the small size of support patches ($i.e$, $48\times 48$), the time and computation cost of conditional feature extraction will be very little compared with BaseNet reconstruction. Without designing new complicated SR network, the proposed framework simply uses 10 res-blocks as BaseNet (called SRResNet-10) and achieves superior performance with blind methods. For complicated degradations, CMDSR even outperforms the non-blind models. It should be noted that our framework has no strict restrictions on the BaseNet structure. The ablation experiments in Table~\ref{other structures} demonstrates that CMDSR can be extended to other SISR models. To the best of our knowledge, the proposed CMDSR is the first hyper-network framework for blind SISR with multiple degradations.

In summary, our overall contribution is three-fold:
\begin{enumerate}
\item We present the first blind hyper-network framework to adaptively handle distribution shifts for the SISR task with multiple degradations in task-level.
\item We propose an unsupervised task contrastive loss to extract more discriminative task-level features.
\item Our proposed framework is hot-pluggable, efficient and flexible. We can replace the modules with stronger models for better performance. Hence, our framework can be applied as a general super-resolution framework.
\end{enumerate}

\section{Related Work}

\textbf{Blind Single-Image Super-Resolution.}
Compared with typical SISR models~\cite{dong2015image,kim2016accurate,lim2017enhanced,zhang2018image} which are tailored to specific single downsampler, blind SISR is a more challenging task, which assumes that the blur kernels are unavailable at test time. Previous methods usually combine the well-designed kernel-estimation and typical SISR methods. Michaeli \textit{et al.}~\cite{michaeli2013nonparametric} mined the internal patch recurrence to estimate the blur kernel. Bell \textit{et al.}~\cite{bell2019blind} proposed KernelGAN to learn the blur kernel distribution. In order to relieve the mismatch between the estimated kernel and the real kernel, IKC~\cite{gu2019blind} iteratively trained the estimation and correction networks. Although the accuracy of estimated kernel is largely improved, it remains very challenging for severe degradation. Inspired by zero-shot and self-supervised learning, ZSSR~\cite{shocher2018zero} efficiently exploited the internal recurrence of information inside an image. But this image-specific model requires self-training for each LR image, which is time-comsuming and cannot be applied to deep structure.

\textbf{Multiple Degradations.}
Few attention has been paid to SISR with multiple degradations despite its important for real applications. Based on the perspective from maximum a posteriori (MAP) framework, existing non-blind methods (such as SRMD~\cite{zhang2018learning} and UDVD~\cite{xu2020unified}) concatenate LR image, predefined blur kernel and noise maps as the input. Thus, SR result closely depends on both the LR image and degradation patterns. The blind schemes~\cite{zhang2018learning,maeda2020unpaired} are usually based on the sequential combinations of denoising~\cite{zhang2017beyond}, blur estimation~\cite{bell2019blind} and SR models~\cite{shocher2018zero}. Besides, CBSR~\cite{liu2020learning} adapted a cascaded architecture, which can be jointly end-to-end learned from training data. All these methods will degrade for distribution shifts. Recently, the unpaired SISR methods~\cite{bulat2018learn, maeda2020unpaired} conducted the domain transfer between the clean and real degraded domains. But it remains challenging to train a stable model for various shifts.

\textbf{Meta-Learning.}
Meta-learning, commonly known as learning how to learn, refers to the process of improving a
learning algorithm over multiple learning episodes. As pointed out by~\cite{lee2018gradient, yao2019automated}, diverse meta-learning methods can be categorized into three groups: (1) Metric based methods~\cite{vinyals2016matching, snell2017prototypical, sung2018learning} perform non-parametric learning in the metric space, which are largely restricted to the popular; (2) Optimization based methods~\cite{andrychowicz2016learning, finn2017model,lee2018gradient,sun2019meta, zhang2020adaptive} use gradient descent to solve the optimization problem of meta-learner. A most famous example is MAML~\cite{finn2017model}, which learns the transferable initial parameters, such that few gradient updates lead to performance improvement. And MTL~\cite{sun2019meta} that leverages the advantages of transfer and meta learning. Recently, ~\cite{zhang2020adaptive} proposes the Adaptive Risk Minimization to handle group distribution shift for image classification; (3) Network based methods~\cite{santoro2016meta,munkhdalai2017meta,oreshkin2018tadam} that use network to learn across tasks and rapidly updates its parameters to a new task.

There are few explorations of meta-learning for SISR. Recently, two gradient based meta-learning models have been proposed. They are the MZSR~\cite{soh2020meta} and MLSR~\cite{park2020fast}, which both employed the typical MAML framework~\cite{finn2017model} to accelerate the self-supervised training. Nevertheless, for large scale factors, the size of self-downsampled LRs on image becomes too small to provide enough information, which which limits their applications to real scenarios.

\section{Proposed Method}
\label{Proposed Method}
\subsection{CMDSR Setting}
Our work focuses on blind SISR with multiple degradations, including blur, noise and down-sampling, which may simultaneously happen in a real-world case~\cite{yang2014single}. The degradation process is formulated as:
\begin{align}
  \label{degradation}
  \textbf{I}_{LR} = (\textbf{I}_{HR}\otimes \textbf{k})\downarrow_s + \textbf{n},
\end{align}
where $\textbf{I}_{LR}$, $\textbf{I}_{HR}$, $\textbf{k}$, $\otimes$, $\downarrow_s$ and $\textbf{n}$ respectively denote LR, HR image, blur kernel, convolution, decimation with scaling factor of $s$. and Gaussian noise. In this paper, we use the configuration in Eq.~(\ref{degradation}). to synthesize the LR images for training.

The key goal of our work is to develop a framework that can adapt and generalize in the face of degradation shift using only a small number of examples. To accomplish this, we need to find the representation that can describe the degradation prior of the LR image and guide the model to adapt to this degradation pattern. As explained in Section~\ref{Introduction}, we observe that the LR images from the same task are degraded with the same pattern. This inspires us to deal with this problem in task-level, rather than in image-level. Therefore, we present a new approach to mine the implicit task-level semantics with different tasks. The extracted feature can be further used as a context prior to adapt the parameters of SR model.

In our framework, we provide two settings to access the training data: (1) The training data should be grouped into different tasks. We consider the multi-degradation distribution $p(\mathcal{T})$ over training tasks $\{\mathcal{T}_1, \mathcal{T}_2, ...\}$. For task $\mathcal{T}_i$, it consists of LR-HR pairs, where LR images $\{x_1^{(i)},...,x_{m}^{(i)}\}$
are synthesized from HR images $\{y_1^{(i)},...,y_{m}^{(i)}\}$ with $i$th degradation configuration. (2) ConditionNet extracts task-level feature from n LR patches (named support set) belonging to the same task, $X_i=\{x_1^{(i)},...,x_{n}^{(i)}\}$ and BaseNet restores the single LR input $x_j^{(i)}$. With these settings, our framework can treat the training data at task-level.

\subsection{Networks of CMDSR}
Our framework consists of ConditionNet and BaseNet, which are shown in Fig.~\ref{CMDSR}. It should be noted that our framework has no strict restrictions on the BaseNet structure. In this paper, the backbone of BaseNet is simply designed as SRResNet-10, which consists of 10 res-blocks.

First, ConditionNet, denoted as $F_c$, extracts the conditional feature $f_c^{(i)}$, which describes the degradation pattern $i$ of the input support set $X^{(i)}$. It is formulated as:
\begin{align}
  f_c^{(i)} = F_c( X^{(i)};\phi) = F_c(x_1^{(i)},...,x_{n}^{(i)};\phi),
\end{align}
where $\phi$ is the parameters of ConditionNet and $n$ denotes the size of support size at each step.
In order to extract the task-level feature, we design a shallow ConditionNet with 2 average pooling layers and 4 convolution layers followed with ReLU and keep the input sample size unchanged during the training and test phrase. The internal channels of convolution layers are $c_{in}*n$, 64, 64, 128, 128.

Then, BaseNet, denoted as $F_{sr}$, adapts its original parameters $\theta$ to $\theta'$ with the conditional feature $f_c^{(i)}$. Since our model aims to be a general framework, we use depth-wise scaling to adapt convolutions without changing the CNN structures. Specifically, we adapt the parameters of 20 conv-layers of internal 10 res-blocks. We use 20 full-connected layers to generate adaptive coefficients with $f_c^{(i)}$ as input. The FC modulation layers change the number of $f_c^{(i)}$ channels to match convolution weights and also adjusts $f_c^{(i)}$ for each conv-layer. Then, the modulated features multiply with the weight of convolution in depth-wise manner:
\begin{align}
  w'_{pq} = w_{pq}*f_{pq}^{'(i)},
\end{align}
where $w_{pq}$ and $w'_{pq}$ are the original and adapted weights, $f_{pq}^{'(i)}$ is the modulated variable corresponding to the $q$th channel of $p$th conv-layer.
Finally, the adapted BaseNet restores the input LR image $x^{(i)}_j$ to the SR image $\hat y^{(i)}_j$. The whole process of BaseNet is formulated as:
\begin{align}
  \hat y^{(i)}_j = F_{sr}(x^{(i)}_j,f_c^{(i)};\theta) = F_{sr}(x^{(i)}_j;\theta').
\end{align}

\subsection{Loss Functions}
\label{Species of Loss Functions}
Owing to the fact that ConditionNet and BaseNet serve different purposes, they have different sensitivity to learning rate and loss functions. Hence, we optimize them alternately with different learning rates and optimization objectives. ConditionNet is trained after every $t$ steps of BaseNet training. The details of loss functions are listed as follows.

\textbf{Reconstruction Loss for BaseNet.} Similar to most SISR models~\cite{kim2016accurate,kim2016deeply,lim2017enhanced,zhang2018image}, we adopt a supervised reconstruction loss to calculate the $L_1$ distance between HR image and output SR image of BaseNet in pixel-wise,
\begin{align}
  \label{reconstruction loss}
  L_{res}= \mathbb{E}_{(\textbf{I}_{HR},\textbf{I}_{LR})\sim p(\mathcal{T})} \parallel \textbf{I}_{HR} - F_{sr}(\textbf{I}_{LR};\theta) \parallel_1.
\end{align}

\textbf{Combined Unsupervised Task Contrastive Loss and Supervised Reconstruction Loss for ConditionNet.}
As the prior knowledge explained before, our ConditionNet should output the conditional features, which are similar to those from the same degradation and dissimilar to others from different degradations. Instead of matching an input to a fixed target, recent works of contrastive learning~\cite{hadsell2006dimensionality,he2020momentum} measure the similarities of sample pairs in a representation space. Inspired by them, we propose a task contrastive loss, which shortens the inner-task distance and enlarges the cross-task distance between different conditional features.

For the inner-task loss, we sample two support sets from the same task, each containing $n$ LR patches, represented as $X_{i}, X'_i$. And ConditionNet $F_c(\cdot$ ;$\theta)$ extracts features $f_{c_1}^{(i)}$, $f_{c_2}^{(i)}$ from $X_{i}$, $X'_i$. The inner-task loss is calculated as:
\begin{align}
  \label{inner-task loss}
  L_{inner}&=\mathbb{E}_{X_i,X'_i\sim p_x(\mathcal{T}_i)}\parallel F_c(X_i;\theta) - F_c(X'_i;\theta) \parallel^2 \notag\\
  &=\mathbb{E}_{X_i,X'_i\sim p_x(\mathcal{T}_i)}\parallel f_{c_1}^{(i)} - f_{c_2}^{(i)} \parallel^2.
\end{align}
For the cross-task loss, we resample $n$ LR images from another task, denoted as support set $X_j$, which show different degradation distribution from $X_i$. Also, ConditionNet $F_c(\cdot$ ;$\theta)$ extracts conditional features $f_{c}^{(i)}$, $f_{c}^{(j)}$ from $X_{i}$, $X_{j}$. Then, the cross-task loss can be calculated as
\begin{align}
  \label{cross-task loss}
  L_{cross}&=\mathbb{E}_{X_i\sim p_x(\mathcal{T}_i), X_j\sim p_x(\mathcal{T}_j), i\neq j}\parallel F_c(X_i;\theta) - F_c(X_j;\theta) \parallel^2 \notag\\
  &=\mathbb{E}_{X_i\sim p_x(\mathcal{T}_i), X_j\sim p_x(\mathcal{T}_j), i\neq j}\parallel f_{c}^{(i)} - f_{c}^{(j)} \parallel^2.
\end{align}
Finally, we use the Logarithm and Exponential transformations to combine $L_{inner}$ and $L_{cross}$. These transformations can smoothly optimize ConditionNet to shorten the inner-task distance and enlarge the cross-task distance. When $L_{inner}$ is small and $L_{cross}$ is large, the combined $L_{con}$ will be close to $0$. The task contrastive loss is formulated as:
\begin{align}
  \label{task contrastive loss}
  L_{con} = \ln (1+e^{-L_{cross}})+\ln (1+e^{L_{inner}}).
\end{align}

During experiments, we find that if we only train ConditionNet by the task contrastive loss in an unsupervised way, the output feature may not be entirely beneficial for the generalization of SISR (as shown in Table~\ref{Combination of Loss Functions.}). To make a balance between task-level feature extraction and SR reconstruction, we combine the reconstruction loss $L_{res}$ in Eq.~(\ref{reconstruction loss}) and task contrastive loss $L_{con}$ in Eq.~(\ref{task contrastive loss}) with coefficient $\lambda$ to constraint ConditionNet, which is formulated as:
\begin{align}
  \label{combined loss}
  L_{com}=L_{con}+\lambda * L_{res}
\end{align}

\subsection{CMDSR Algorithm}
\begin{algorithm}[h]
\label{CMDSR Training}
\caption{CMDSR Training}
\LinesNumbered
\KwData{Distribution over tasks: $p(\mathcal{T})$}
\KwIn{ConditionNet and BaseNet parameters: $\phi$, $\theta$}
\KwIn{Task size: $k$, support size: $n$, update step: $t_0$, loss coefficient: $\lambda$, learning rates: $\alpha$, $\beta$}
\For{$t = 1, 2, ...$}{
Randomly sample $k$ tasks $\mathcal{T}_i\sim p(\mathcal{T})$\;
\ForEach{$\mathcal{T}_i$}{
Random crop $n$ LR-HR patches from $n$ LR-HR images of $\mathcal{T}_i$ as $(X_i, Y_i)\sim p(\mathcal{T}_i)$\;
Use all $n$ LR patches of $X_i$ as the support set to
extract conditional feature $f_c^{(i)}$: $f_c^{(i)}=F_{c}(X_i;\phi)=F_{c}(x_{i_1},...,x_{i_{n}};\phi)$\;
Compute adapted parameters $\theta_i'$ of BaseNet: $F_{sr}(\cdot$ $;\theta_i')=F_{sr}(f_c^{(i)};\theta)$\;
Evaluate the reconstruction loss in Eq.~(\ref{reconstruction loss}): $L_{res}^{(i)}=\sum_{j=1}^n L_1(F_{sr}(x_{i_j};\theta_i'), y_{i_j})$\;
}
{Update BaseNet with reconstruction loss: $\theta\leftarrow\theta-\alpha\nabla_{\theta} \sum_{i=1}^k L_{res}^{(i)}$\;}
\If{$(t$ Mod $t_0)$ $=0$}{
Resample $n$ LR patches for $k$ support sets from $k$ tasks $\mathcal{T}_i$ of line 4: $X'_i\sim p_x(\mathcal{T}_i)$\;
Resample $n$ LR patches for $k$ support sets from another $k$ tasks $\mathcal{T}_j$: $X_j\sim p_x(\mathcal{T}_j), j\neq i$\;
\ForEach{$\mathcal{T}_i,\mathcal{T}_j,j\neq i$}{
Evaluate inner-task loss as Eq.~(\ref{inner-task loss}):
$L_{inner}^{(i)}=L_2(F_c(X_{i};\theta),F_c(X'_i;\theta))$\;
Evaluate cross-task loss as Eq.~(\ref{cross-task loss}):
$L_{cross}^{(i)}=L_2(F_c(X_i;\theta),F_c(X_j;\theta))$\;
Evaluate task contrastive loss as Eq.~(\ref{task contrastive loss}):
$L_{con}^{(i)}=L_{con}(L_{inner}^{(i)},L_{cross}^{(i)})$\;
}
{Update ConditionNet with combined loss: $\phi\leftarrow\phi-\beta\nabla_{\phi} \sum_{i=1}^k (L_{con}^{(i)}+\lambda L_{res}^{(i)})$\;}
}
}
\end{algorithm}

\textbf{Training Stage.} Algorithm~\ref{CMDSR Training} shows the training procedure of CMDSR model. ConditionNet and BaseNet are alternately trained until they converge. In line 4, $k$ tasks are randomly sampled from degradation distribution $p(\mathcal{T})$ for each step. In Line 3-9, BaseNet is adapted and supervised with HR-LR pairs. In Line 10-18, for every $t_0$ steps, ConditionNet is optimized with the combined unsupervised $L_{con}$ of Line 16 and supervised $L_{res}$ of line 7.

\begin{algorithm}[htbp]
\label{CMDSR Test}
\caption{CMDSR Testing}
\LinesNumbered
\KwData{LR test image: $\textbf{I}_{LR}$}
% , LR self-support set from LR image: $X$}
\KwIn{Trained parameters of ConditionNet $F_c$ and BaseNet $F_{sr}$: $\phi$, $\theta$}
\KwOut{Restored SR image: $\textbf{I}_{SR}$}
Randomly crop $n$ patches of $I_{LR}$ as the support set $X$\;
Extract conditional feature $f_c=F_{c}(X;\phi)$\;
Compute adapted parameters $\theta'$ of BaseNet: \quad\quad\quad
$F_{sr}(\cdot$ $;\theta')=F_{sr}(f_c;\theta)$\;
\Return $\textbf{I}_{SR}=F_{sr}(\textbf{I}_{LR};\theta')$
\end{algorithm}

\textbf{Testing Stage.} Algorithm~\ref{CMDSR Test} shows the training procedure of CMDSR model. For test support set $X$, we can randomly sample patches from other LR images, which have the same degradation pattern with $\textbf{I}_{LR}$, or from $\textbf{I}_{LR}$ itself.
% For convenience, we choose the self-patches to get the support set at test time.
For unknown LR in real scenario, it's hard to acquire other images which have the same degradation pattern. Therefore, we randomly crop patches from LR itself (self-patches) to get the support set at test time in all our experiments. With the conditional feature extracted from the support set, BaseNet performs fast adaptation to the test distribution in one step and produces the restored SR image.

\section{Experiments}

% \vspace{-0.1cm}
\subsection{Experimental Setting}
\label{Experimental Setups}
As introduced before, the input of CMDSR consists of two parts: the support set for the ConditionNet and the LR image for the BaseNet. When training, n random patches are cropped from n randomly sampled LRs of the same task and this process is conducted for k training tasks. Hence, during training, the sizes of support sets and LR images are set as  $[k,n*3,h,w]$ and $[k,3,h,w]$ separately,  where size 3 indicates RGB channels. When testing, the full LR image $[1,3,H,W]$ is fed into BaseNet and n random patches are cropped from LR itself to compose the self-support set $[1,n*3,h,w]$ for testing. All uniform random croppings allow for overlapping. In training configurations, we set the task size as $k=8$ and the size of support set $n$ to 20. The patch size $h \times w$ is $48 \times 48$. The update step $t_0$ is $10$, which means that the ConditionNet is joined for training after the BaseNet has been trained for 9 steps. The loss coefficient $\lambda$ of Eq.~(\ref{combined loss}) is $0.1$. The initial learning rates
$\alpha,\beta$ of the BaseNet and ConditionNet are set to $10^{-3}$ and $10^{-4}$. The ADAM optimizer~\cite{kingma2015adam} is applied.

\begin{table*}[t]
  \centering
  \caption{Average PSNR values with scale factor $\times$4 on \textit{Simple}, \textit{Middle} and \textit{Severe} degradations. We use the provided opensource codes of SOTA models to compute their results, except results of IRCNN~\cite{zhang2017learning} and UDVD~\cite{xu2020unified}, which are directly extracted from their publications. Best and second best results are highlighted in \textcolor{red}{red} and \textcolor{blue}{blue}.
  }
  \label{Matched Degradation}
  \centering
  \begin{tabular}{cccccccc}
  \hline
  \multirow{2}{*}{Degradation}&\multirow{2}{*}{Kernel}&\multirow{2}{*}{Noise}&\multirow{2}{*}{Model Type}&\multirow{2}{*}{Models}&\multicolumn{3}{c}{Datasets}\\
  &&&&&Set5&Set14&BSD100\\
  \hline
  \hline
  \multirow{12}{*}{\makecell[c]{(A) \textit{Simple} Degradation\\In-Distribution}}&\multirow{13}{*}{$\sigma_G^{0.2}$}&\multirow{13}{*}{$\sigma_N^{15}$}&BI-structured SR model&RCAN+~\cite{zhang2018image}& 24.90& 23.87& 23.42\\
  \cline{4-8}
  & & & \multirow{2}{*}{\makecell[c]{Blind multi-degraded\\ SR model}}&ZSSR~\cite{shocher2018zero}& 25.40& 24.30& 24.05\\
  & & & &IRCNN~\cite{zhang2017learning}& 28.35& -& -\\
  \cline{4-8}
  & & & \multirow{4}{*}{\makecell[c]{Blind denoising/deblurring\\+ Blind SR model}}&DnCNN~\cite{zhang2017beyond}+KernelGAN~\cite{bell2019blind}+ZSSR~\cite{shocher2018zero}& 27.02& 25.46& 25.34\\
  & & & &DnCNN~\cite{zhang2017beyond}+MZSR~\cite{soh2020meta}& 26.57& 25.10& 24.72\\
  & & & &DnCNN~\cite{zhang2017beyond}+IDN~\cite{hui2018fast}+MLSR~\cite{park2020fast}& 26.32& 25.14& 24.63 \\
  & & & &DnCNN~\cite{zhang2017beyond} + IKC~\cite{gu2019blind}& 28.16& 26.11& 25.68\\
  \cline{4-8}
  & & & \makecell[c]{Blind denoising+\\Gt blur kernel maps+\\Non-blind SR model}&DnCNN~\cite{zhang2017beyond} + SRMDNF~\cite{zhang2018learning}& 28.31& 26.19& 25.79\\
  \cline{4-8}
  & & & \multirow{2}{*}{\makecell[c]{Gt Degradation maps+\\ Non-blind SR model}}&SRMD~\cite{zhang2018learning}& \textcolor{blue}{28.79}& \textcolor{blue}{26.48}& \textcolor{blue}{25.95}\\
  & & & &UDVD~\cite{xu2020unified}& \textcolor{red}{29.04}& \textcolor{red}{26.82}& \textcolor{red}{26.08}\\
  \cline{4-8}
  & & & Conditional hyper-network&Our CMDSR& 28.35& 26.23& 25.83\\
  \hline
  \hline
  \multirow{12}{*}{\makecell[c]{(B) \textit{Middle} Degradation\\In-Distribution}}&\multirow{13}{*}{$\sigma_G^{2.6}$}&\multirow{13}{*}{$\sigma_N^{15}$}&BI-structured SR model&RCAN+~\cite{zhang2018image}& 23.32& 22.54& 22.61\\
  \cline{4-8}
  & & & \multirow{2}{*}{\makecell[c]{Blind multi-degraded\\ SR model}}&ZSSR~\cite{shocher2018zero}& 24.91& 23.74& 23.57\\
  & & & &IRCNN~\cite{zhang2017learning}& 24.36& -& -\\
  \cline{4-8}
  & & & \multirow{4}{*}{\makecell[c]{Blind denoising/deblurring\\+ Blind SR model}}
  &DnCNN~\cite{zhang2017beyond}+MZSR~\cite{soh2020meta}& 25.86& 24.31& 24.40\\
  & & & &DnCNN~\cite{zhang2017beyond}+IDN~\cite{hui2018fast}+MLSR~\cite{park2020fast}& 25.89& 24.44& 24.32 \\
  & & & &DnCNN~\cite{zhang2017beyond}+KernelGAN~\cite{bell2019blind}+ZSSR~\cite{shocher2018zero}& 26.08& 24.66& 24.65\\
  & & & &DnCNN~\cite{zhang2017beyond} + IKC~\cite{gu2019blind}& 26.84& 25.09& \textcolor{blue}{25.02}\\
  \cline{4-8}
  & & & \makecell[c]{Blind denoising+\\Gt blur kernel maps+\\Non-blind SR model}&DnCNN~\cite{zhang2017beyond} + SRMDNF~\cite{zhang2018learning}& 23.85& 21.04& 21.79\\
  \cline{4-8}
  & & & \multirow{2}{*}{\makecell[c]{Gt Degradation maps+\\Non-blind SR model}}&SRMD~\cite{zhang2018learning}& 26.82& 25.12& 24.86\\
  & & & &UDVD~\cite{xu2020unified}& \textcolor{blue}{26.98}& \textcolor{blue}{25.33}& 24.96\\
  \cline{4-8}
  & & & Conditional hyper-network&Our CMDSR& \textcolor{red}{27.10}& \textcolor{red}{25.39}&\textcolor{red}{25.12}\\
  \hline
  \hline
  \multirow{12}{*}{\makecell[c]{(C) \textit{Severe} Degradation\\Out-of-Distribution}}&\multirow{13}{*}{$\sigma_G^{ani}$}&\multirow{13}{*}{$\sigma_N^{50}$}&BI-structured SR model&RCAN+~\cite{zhang2018image}& 16.32& 15.83& 15.81\\% &25.3 &24.29 &23.64 &24.95 &23.92 &23.33\\
  \cline{4-8}
  & & &\makecell[c]{Blind multi-degraded\\ SR model}&ZSSR~\cite{shocher2018zero}& 17.89& 17.46& 17.79\\
  \cline{4-8}
  & & & \multirow{4}{*}{\makecell[c]{Blind denoising/deblurring\\ + Blind SR model}}
  &DnCNN~\cite{zhang2017beyond}+MZSR~\cite{soh2020meta}& 21.73& 20.80& 21.49\\
  & & & &DnCNN~\cite{zhang2017beyond}+IDN~\cite{hui2018fast}+MLSR~\cite{park2020fast}& 21.31& 20.46& 21.32\\
  & & & &DnCNN~\cite{zhang2017beyond}+KernelGAN~\cite{bell2019blind}+ZSSR~\cite{shocher2018zero}& 22.32& 21.69& 22.34\\
  & & & &DnCNN~\cite{zhang2017beyond} + IKC~\cite{gu2019blind}& 22.18& 21.63& 22.23\\
  \cline{4-8}
  & & & \makecell[c]{Blind denoising+\\Gt blur kernel maps+\\Non-blind SR model}&DnCNN~\cite{zhang2017beyond} + SRMDNF~\cite{zhang2018learning}& 21.63& 21.18& 21.99 \\
  \cline{4-8}
  & & &\makecell[c]{Gt degradation maps+\\Non-blind SR model}&SRMD~\cite{zhang2018learning}& \textcolor{blue}{22.43}& \textcolor{blue}{21.83}& \textcolor{blue}{22.43}\\
  \cline{4-8}
  & & & Conditional hyper-network&Our CMDSR& \textcolor{red}{23.07}& \textcolor{red}{22.14}& \textcolor{red}{23.03}\\
  \hline
\end{tabular}
\end{table*}

We use the LR-HR pairs of DIV2K~\cite{agustsson2017ntire} for training. Following previous works~\cite{zhang2018learning,xu2020unified}, the degraded LR images of different tasks are synthesized based on Eq.~(\ref{degradation}). As shown in Eq. 1, the multi-degradation applies a sequence of degrading effects on HR images, following the order of blurring, bicubic downsampling and then adding noise. To ensure consistency with previous works~\cite{zhang2018learning,xu2020unified}, we use $15\times 15$ isotropic Gaussian blur kernels and the Additive White Gaussian Noise (AWGN). The blur kernel widths $\sigma_{G}$ are discrete in range of [0.2, $s$] for scale factor $s$ with a stride of 0.1. For noise, we set the AWGN with the continuous noise levels $\sigma_{N}$ in range [0, 75]. For each iteration, the discrete kernel width and continuous noise level are randomly sampled. The total task number of our training data is infinite. We set the scale factor $s=4$ and all the experiments were conducted on NVIDIA Tesla-V100 GPUs.

It should be noted that previous methods~\cite{zhang2018learning,gu2019blind} choose the isotropic and anisotropic Gaussian blur kernels for generating blur degradation. Since we only use isotropic Gaussian blur kernels, our training degradation patterns ($\sigma_{G}$, $\sigma_{N}$) are the subspace of ($\sigma_{G}$, $\sigma_{G}^{ani}$, $\sigma_{N}$) in SOTAs. Therefore, the training sets of SOTAs can cover all degradation information of ours. Although this is ``unfair" for our model, we want to highlight the ability of our model for degradation shift in this ``unfair" setting and verify the effectiveness of our work.

\begin{figure*}[htbp]
  % \vspace{-0.4cm}
  \centering
  \includegraphics[width=15cm]{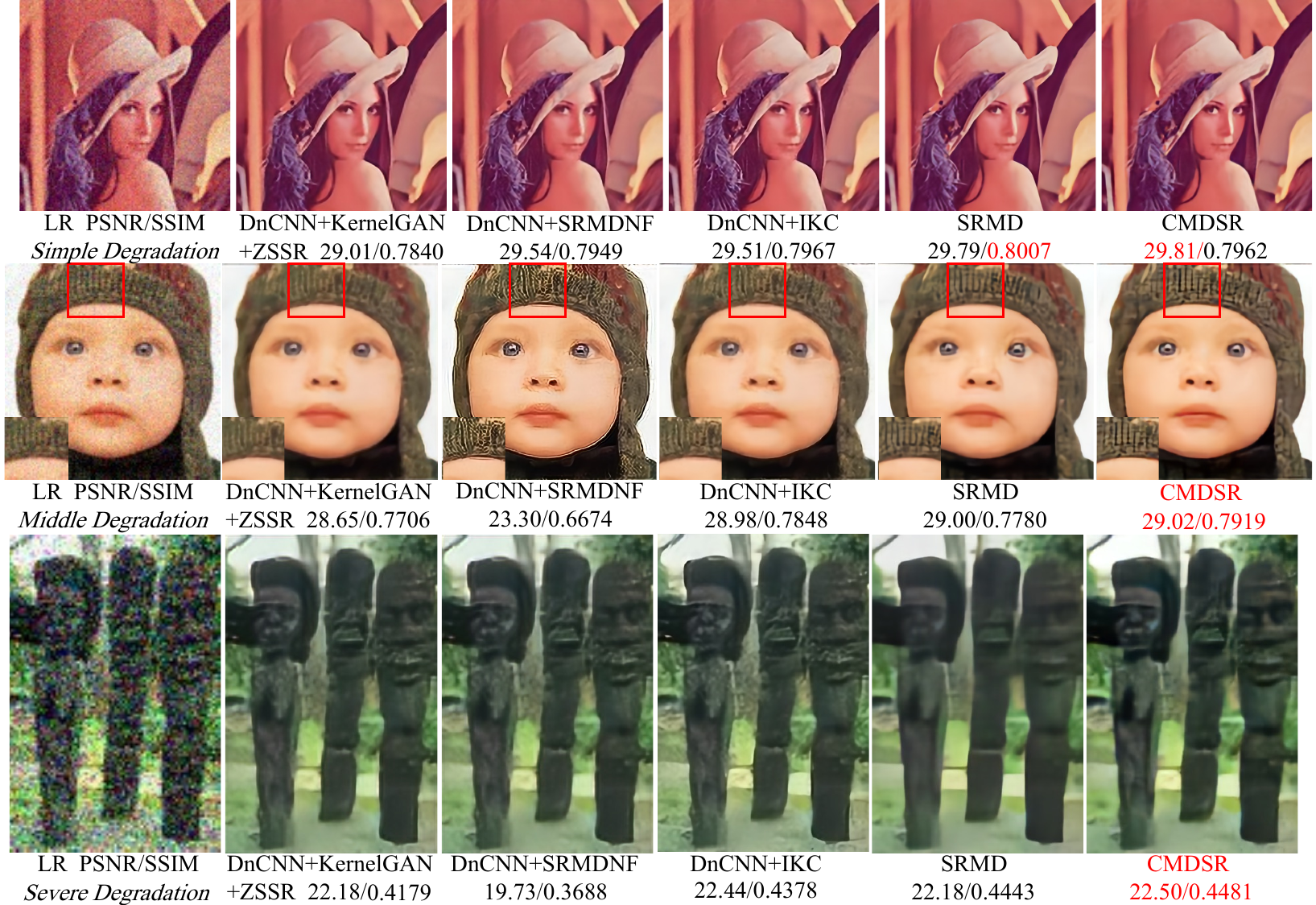}
  \caption{SR perceptual results (×4) of images on \textit{Simple}, \textit{Middle} and \textit{Severe} degradations. Best results are highlighted in \textcolor{red}{red}.  For \textit{Simple} degradation, CMDSR outperforms all the blind methods and is competitive with non-blind method, the SRMD. For \textit{Middle} and \textit{Severe} degradations, our CMDSR even outperforms the non-blind SRMD. Better to enlarge the images for more visual details.}
  \label{visual_anti_50_2}
  % \vspace{-0.5cm}
\end{figure*}

% \vspace{-0.1cm}
\subsection{Experiments on Synthetic Multi-degradated Images}
\label{Experiments on Synthetic Images}
To demonstrate the effectiveness and generalization of our framework for multi-degradations, we evaluate the proposed CMDSR from the perspectives of matched degradation and shift degradation. First, we use the \textit{Simple} and \textit{Middle} testing sets, which are in range of training distribution. Since our framework is trained with isotropic Gaussian blur kernel, we also add the \textit{Severe} testing set with anisotropic Gaussian blur kernel to validate whether CMDSR can handle degradation shift. More precisely, three testing sets are synthesized as:
\begin{enumerate}
\item[(A)] \textit{Simple}: $7 \times 7$ isotropic Gaussian blur kernel with kernel width $\lambda=0.2$ followed by BI downsampler ($\sigma_G^{0.2}$) and AWGN with noise level 15 ($\sigma_N^{15}$).
\item[(B)] \textit{Middle}: $7 \times 7$ isotropic Gaussian blur kernel with kernel width $\lambda=2.6$ followed by BI downsampler ($\sigma_G^{2.6}$) and AWGN with noise level 15 ($\sigma_N^{15}$).
\item[(C)]  \textit{Severe}: $7 \times 7$ anisotropic Gaussian blur kernel with kernel width $\lambda_1=4$, $\lambda_2=1$, angle $\Theta=-0.5$ followed by BI downsampler ($\sigma_G^{ani}$) and AWGN with noise level 50 ($\sigma_N^{50}$).
\end{enumerate}
The \textit{Simple} and \textit{Middle} are in distribution of training data and the \textit{Severe} is out of distribution. We follow~\cite{zhang2018learning,xu2020unified} to set the degradation parameters of three testing sets for fair comparison with other SOTA methods for consistent comparison.

To be specific, we compare the proposed framework with non-blind and blind methods. For non-blind methods, we make comparison with two latest models, SRMD~\cite{zhang2018learning} and UDVD~\cite{xu2020unified}, which utilize the accurate blur kernel and noise maps as the additional inputs. For blind methods, the SOAT BI structured SR model, RCAN~\cite{zhang2018image} is first compared. Since most blind SR methods for multiple degradations have not been studied sufficiently, except ZSSR~\cite{shocher2018zero} (5000 steps) and IRCNN~\cite{zhang2017learning}, we follow~\cite{zhang2018learning, bell2019blind, maeda2020unpaired} to add cascaded schemes, which combine SR models with blind denoising and deblurring methods: DnCNN\cite{zhang2017beyond} + KernelGAN\cite{bell2019blind} + ZSSR\cite{shocher2018zero} (5000 steps), DnCNN~\cite{zhang2017beyond} + IKC~\cite{gu2019blind}. Although two previous meta SR methods~\cite{soh2020meta, park2020fast} have limitations with large scale factor ($\times$4), we still add them as DnCNN~\cite{zhang2017beyond} + MZSR~\cite{soh2020meta} (10 steps), DnCNN~\cite{zhang2017beyond} + IDN~\cite{hui2018fast} + MLSR~\cite{park2020fast} (10 steps). To evaluate the mutual negative influence between cascaded stages, we also add a baseline by combining blind denoiser and non-blind SR model, DnCNN~\cite{zhang2017beyond} + SRMDNF~\cite{zhang2018learning}. %Due to the page limit, we present PSNR results of $\times 4$ SR tasks. More results are listed in \textit{Supplementary}.
\begin{figure*}[t]
  \includegraphics[width=14.2cm]{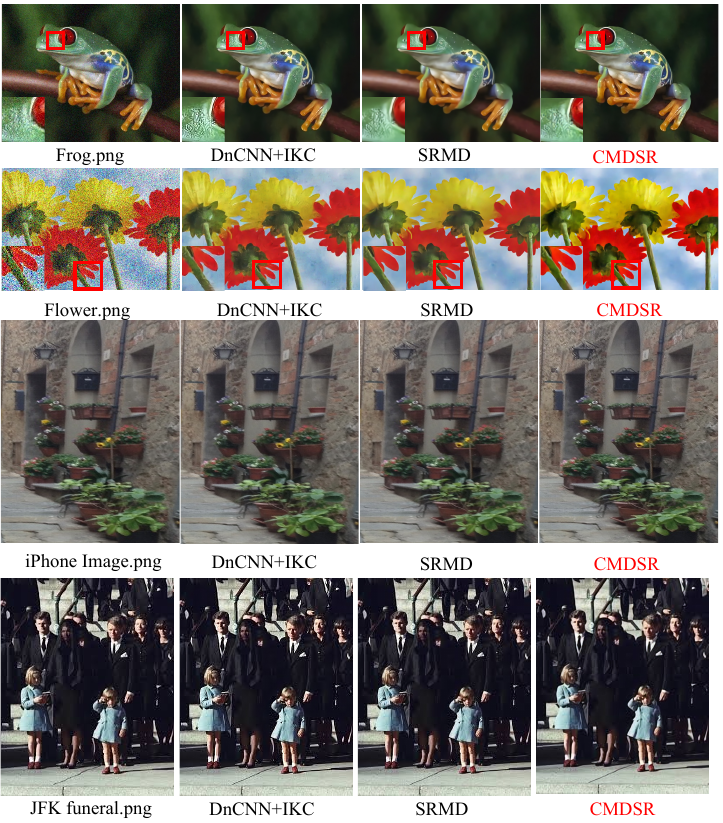}
  \centering
  \caption{SR perceptual results (×4) of real-captured images. Best results are highlighted in \textcolor{red}{red}. Our CMDSR produces best results with \textbf{less artifacts and brighter color}. Better to enlarge the images for more visual details.}
  \label{Real_img}
\end{figure*}

% \textbf{Degradation-in-Distribution.}
\subsubsection{Degradation-in-Distribution}
The Section A and B of Table~\ref{Matched Degradation} show the PSNR values on \textit{Simple} and \textit{Middle} degradations, where degradation patterns match the range of training data. Due to the unawareness of multiple degradations, the BI structured model RCAN~\cite{zhang2018image} produces the worse PSNR. When kernel width and noise level are increasing, the cascaded blind methods suffer the mutual negative influence between different stages. This is because the denoiser will make the LR image more blurred and lead to kernel mismatch, which increases the difficulty of deblurring. The severe performance drops of PSNR as shown in Table~\ref{Matched Degradation} and the over-sharp results of DnCNN~\cite{zhang2017beyond} + SRMDNF~\cite{zhang2018learning} in Fig.~\ref{visual_anti_50_2} clearly illustrates this phenomenon. Our CMDSR achieves better PSNR than all blind schemes for \textit{Simple} degradation, but a little lower than the non-blind methods, SRMD~\cite{zhang2018learning} and UDVD~\cite{xu2020unified}, because they take the accurate blur kernel and noise maps as the additional inputs. However, it is noted that when degradation is more complicated, the generalization of our adaptive framework becomes prominent. Our CMDSR achieves the best performance for \textit{Middle} degradation, even better than the non-blind methods. As shown in Fig.~\ref{visual_anti_50_2}, CMDSR produces sharper and clearer SR results. These results demonstrate that CMDSR is an effective blind framework for multiple degradations.

\subsubsection{Degradation-Out-of-Distribution}
The results in Section C of Table~\ref{Matched Degradation} show PSNR values on \textit{Severe} degradation, where the degradation levels are higher and the blur kernel is out of the distribution of training data. The qualitative comparisons are presented in Fig.~\ref{visual_anti_50} and Fig.~\ref{visual_anti_50_2}. Our CMDSR significantly outperforms all the blind and the non-blind methods, because the parameters of BaseNet are not fixed but adaptive for a new degradation when testing. It should be emphasized that non-blind SRMD~\cite{zhang2018learning} is trained with both isotropic and anisotropic Gaussian blur kernel, while our blind CMDSR  trained with isotropic blur kernels still achieves better qualitative and quantitative results. These results further demonstrate the generalization of our CMDSR to handle distribution shifts.

\begin{table}[t]
  \centering
  \caption{Average PSNR values with scale factor $\times$4 on noise-free degradations. Best and second best results are highlighted in \textcolor{red}{red} and \textcolor{blue}{blue}. }
  \label{noise_free}
  \centering
  \begin{tabular}{ccccc}
  \hline
  \multirow{2}{*}{Models}&\multirow{2}{*}{Kernel}&\multicolumn{3}{c}{Datasets}\\
  &&Set5&Set14&BSD100\\
  \hline
  ZSSR&\multirow{5}{*}{$\sigma_G^{0.2}$}& 28.87& 27.15& 26.68\\
  %IRCNN& & 31.02& -& -\\
  SRMDNF& & 31.96& 28.35& 27.49\\
  IKC& & \textcolor{red}{32.39}& \textcolor{blue}{28.77}& 27.58\\
  UDVD& & \textcolor{blue}{32.31}& \textcolor{red}{28.78}& \textcolor{red}{27.70}\\
  CMDSR & & 31.82& 28.53& \textcolor{blue}{27.60} \\
 \hline
  ZSSR&\multirow{5}{*}{$\sigma_G^{2.6}$}& 27.69& 26.06& 25.92\\
  %IRCNN& & 30.06& -& -\\
  SRMDNF& & 31.77& 28.26& 27.43\\
  IKC& & \textcolor{red}{32.05}& \textcolor{red}{28.55}& 27.47\\
  UDVD& & \textcolor{blue}{31.99}& \textcolor{red}{28.55}& \textcolor{red}{27.55}\\
  CMDSR & &31.79 & \textcolor{blue}{28.36}& \textcolor{blue}{27.50}\\
  \hline
    ZSSR&\multirow{4}{*}{$\sigma_G^{ani}$}& 26.24& 25.33& 25.02\\
  SRMDNF& & \textcolor{blue}{30.48}& 27.23& \textcolor{blue}{26.85} \\
  IKC& & \textcolor{red}{30.51}& \textcolor{red}{27.31}& 26.70\\
  CMDSR & &29.94 & \textcolor{blue}{27.29}& \textcolor{red}{26.95}\\
  \hline
\end{tabular}
\end{table}

\subsection{Experiments on Synthetic Noise-free Images}
As we have explained before, although existing methods can perform well on single degradation (\textit{eg.} noise-free), they are difficult to deal with multiple degradations, and the multiple degradations are more complicated in real tasks. However, it is still a valid assessment on noise-free degradation. Here, we present the experimental results on noise-free degradation in Table~\ref{noise_free}. Compare to well-designed SOTAs, such as blind IKC and non-blind UDVD, our CMDSR achieves little worse performance. When degradation is more complicated and out of distribution, the performance of our CMDSR can get improved, which also proves the adaptation of our model on noise-free degradation. It should be noted that since different degradation patterns are mixed together, it's a tough problem for existing blind SR models to conduct accurate degradation estimation on multi-degradations. Therefore, the superiority of adapted our CMDSR are more significant on multi-degraded images.

% \vspace{-0.1cm}
\subsection{Perceptual Visualization on Real Images}
We further extend the experiments to real images. The most representative blind and non-blind methods, DnCNN~\cite{zhang2017beyond} + IKC~\cite{gu2019blind} and SRMD~\cite{zhang2018learning} are compared with our framework. Since there are no GT degradation patterns for real images, SRMD~\cite{zhang2018learning} is searched by manual grid as in~\cite{zhang2018learning}.
The qualitative results of real images~\cite{lebrun2015noise,zhang2018learning} are shown in Fig.~\ref{Real_img}. It is clear that the blind scheme produces over-sharp results and non-blind SRMD~\cite{zhang2018learning} fails to recover sharp edges. Overall, CMDSR produces the best results with less artifacts, sharper edges, and even brighter color.

% \vspace{-0.1cm}
\subsection{Ablation Experiments}
\label{Ablation Experiments}
In this section, we present the systematic ablation studies to explain the important implementation details and validate the effectiveness of our framework. For fair comparison, all the ablation studies were conducted with training data and settings as described in Section~\ref{Experimental Setups}.

\subsubsection{Computational and Parameter Comparison}
We present the SOTA blind and non-blind SR models, RCAN~\cite{zhang2018image}, ZSSR~\cite{shocher2018zero}, DnCNN\cite{zhang2017beyond} + KernelGAN\cite{bell2019blind} + ZSSR\cite{shocher2018zero} (5000 steps), DnCNN~\cite{zhang2017beyond} + IKC~\cite{gu2019blind}, SRMD~\cite{zhang2018learning}, UDVD\footnote{The codes of UDVD~\cite{xu2020unified} have not been released. We reconstruct the backbone of UDVD according to its original paper to calculate the model parameters (Pytorch version) and average inference time.}~\cite{xu2020unified}, SRResNet-10 and CMDSR to conduct experiments. The model parameters, average inference time and PSNR results for $\times 4$ SISR of Set5 with \textit{Middle} Degradation are listed in Table~\ref{Computational and Parameter Comparison}. It is clear that our proposed CMDSR can achieve better results with fast speed and relatively less parameters.

\begin{table*}[t]
  % \vspace{-1.1cm}
  % \setlength{\belowcaptionskip}{-0.7cm}
% \begin{center}
\centering
\caption{Comparisons on model parameters, average inference time and PSNR results for $\times 4$ SISR of Set5 with \textit{Middle} degradation. It is clear that our proposed CMDSR can achieve better results with fast speed and relatively less parameters.}
\label{Computational and Parameter Comparison}
\centering
\begin{tabular}{ccccc}
\hline
Model&RCAN+~\cite{zhang2018image}&ZSSR~\cite{shocher2018zero} (5000 steps) &\makecell[c]{DnCNN\cite{zhang2017beyond} + KernelGAN\cite{bell2019blind}\\ + ZSSR\cite{shocher2018zero} (5000 steps)} &\makecell[c]{DnCNN~\cite{zhang2017beyond} + IKC~\cite{gu2019blind}\\ (10 iterations)} \\
\hline
% ZSSR0.23M DnCNN 0.67M IKC 1.49M KernelGAN 0.18M
% DnCNN 20ms
Params.&15.6M&0.23M&1.08M&2.16M\\
Sec.& 507.32ms& 183.85s& 213.93s& 780.40ms\\
PSNR& 23.32& 24.91& 26.08&26.84 \\
\hline
\hline
Model&SRMD~\cite{zhang2018learning}&UDVD~\cite{xu2020unified}&SRResNet-10&\makecell[c]{CMDSR\\ (SRResNet-10 + ConditionNet)}\\
\hline
Params.&1.55M&4.83M&1.04M&1.48M\\
Sec.& 6.28ms& 42.72ms& 29.84ms& 39.19ms\\
PSNR& 26.82& 26.98& 26.09& \textcolor{red}{27.10}\\
\hline
\end{tabular}
% \end{center}
\end{table*}

\subsubsection{BaseNet w/ and w/o ConditionNet} We first evaluate the performance of BaseNet with and without ConditionNet to show the importance of conditional feature. Although ConditionNet is not directly used for SISR, it involves more parameters. For a fair comparison, we add SRResNet-16, where the number of parameters nearly equals to the completed CMDSR. The SRResNet-16 is then trained with the same synthetic data as ours. As shown in Table~\ref{BaseNet w/ and w/o ConditionNet.}, the PSNR result of CMDSR is much better, which demonstrates the significance of conditional hyper-network for unsupervised task-level feature extraction.

\begin{table}[htbp]
\centering
\caption{Average $\times$4 PSNR of SRResNet-10, SRResNet-16 and SRResNet-10+ConditionNet on Set5 with \textit{Middle} degradation.}
\label{BaseNet w/ and w/o ConditionNet.}
\centering
\begin{tabular}{ccc}
\hline
Model&Parameters&PSNR\\
\hline
\hline
SRResNet-10 w/o ConditionNet&1.04M&26.09\\
SRResNet-16 w/o ConditionNet&1.48M&26.62\\
SRResNet-10 w/ ConditionNet (Our CMDSR)&1.46M&\textcolor{red}{27.10}\\
\hline
\end{tabular}
% \vspace{-0.6cm}
\end{table}

\subsubsection{Visualizations of Conditional Features}
\begin{figure}[h]
  % \vspace{-0.5cm}
  \centering
  \includegraphics[width=5cm]{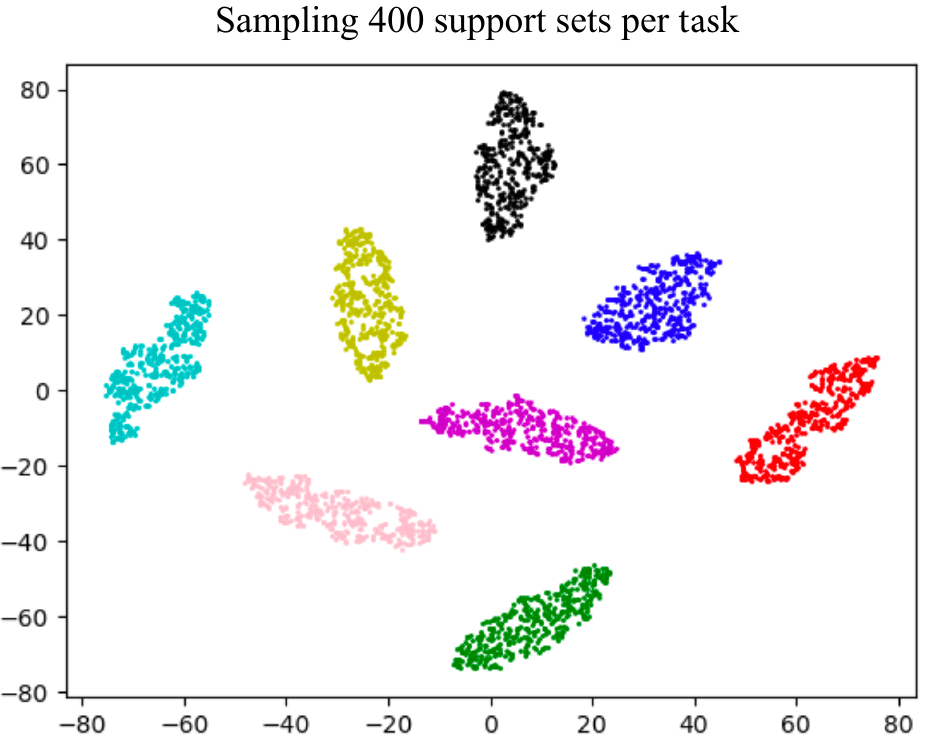}
  \caption{The t-SNE visualization of modulated conditional features. Different colors represent different tasks.}
  \label{visual_conditional_feature}
\end{figure}
To show ConditionNet can efficiently extract task-level features, we compare the conditional features between inner-tasks and cross-tasks. Using DIV2K validation set, we randomly sample 8 different tasks and sample 400 support sets for each task. We choose the conditional features modulated by the first modulation layer and show the t-SNE~\cite{van2008visualizing} visualizations in Fig.~\ref{visual_conditional_feature}.
% We also list the values of 64 channels for each feature in Table~\ref{cross-task conditional features} and Table~\ref{inner-task conditional features}.
It is clear that the modulated features of inner-tasks are similar and those from cross-tasks are significantly different, which is consistent with the prior knowledge in Fig.~\ref{Task-level}.

\begin{figure}[htbp]
  % \vspace{-0.5cm}
  \centering
  \includegraphics[width=7cm]{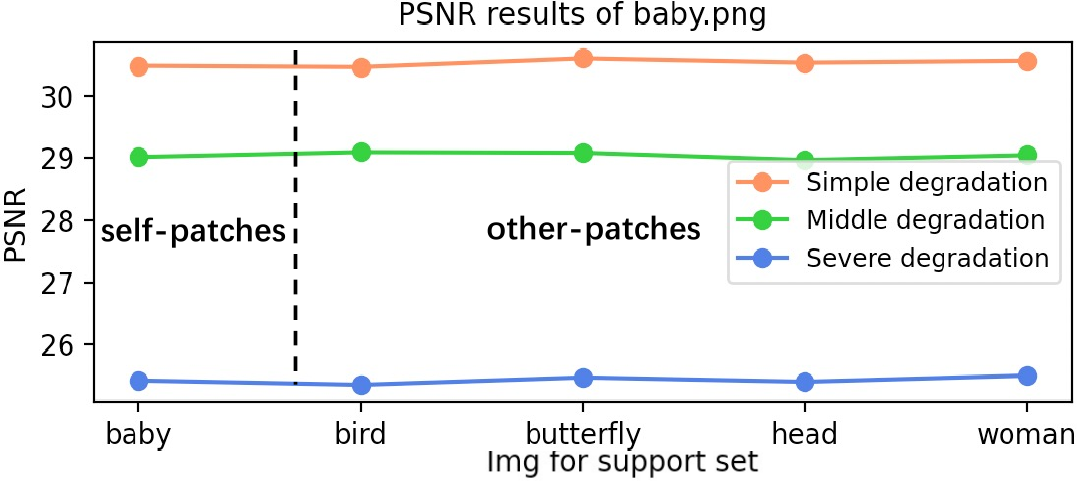}
  \caption{PSNR results of \textit{baby}.png using different support patches of Set5.}
  \label{support_patch}
\end{figure}
\subsubsection{Stability with Different Support Sets}
Our framework aims to extract the common degradation patterns from different LR images of the same task. Therefore, the conditional features should be content-invariant. To demonstrate this, we conduct experiments to evaluate the stability of SR results with different support sets. As shown in Fig.~\ref{support_patch}, we evaluated the PSNR values of \textit{baby}.png using different support patches of Set5. We randomly cropped the self-support-patches from  \textit{baby}.png itself and other-support-patches from other LR images of Set5, separately. The self-support-patches and other-support-patches from the same task share the same degradation but have different contents. As shown in Fig.~\ref{support_patch}, it is clear that our framework achieves relatively stable performance. This further verifies that our model can extract degradation prior from different image contents at task-level.

\begin{figure}[h]
  \includegraphics[width=5cm]{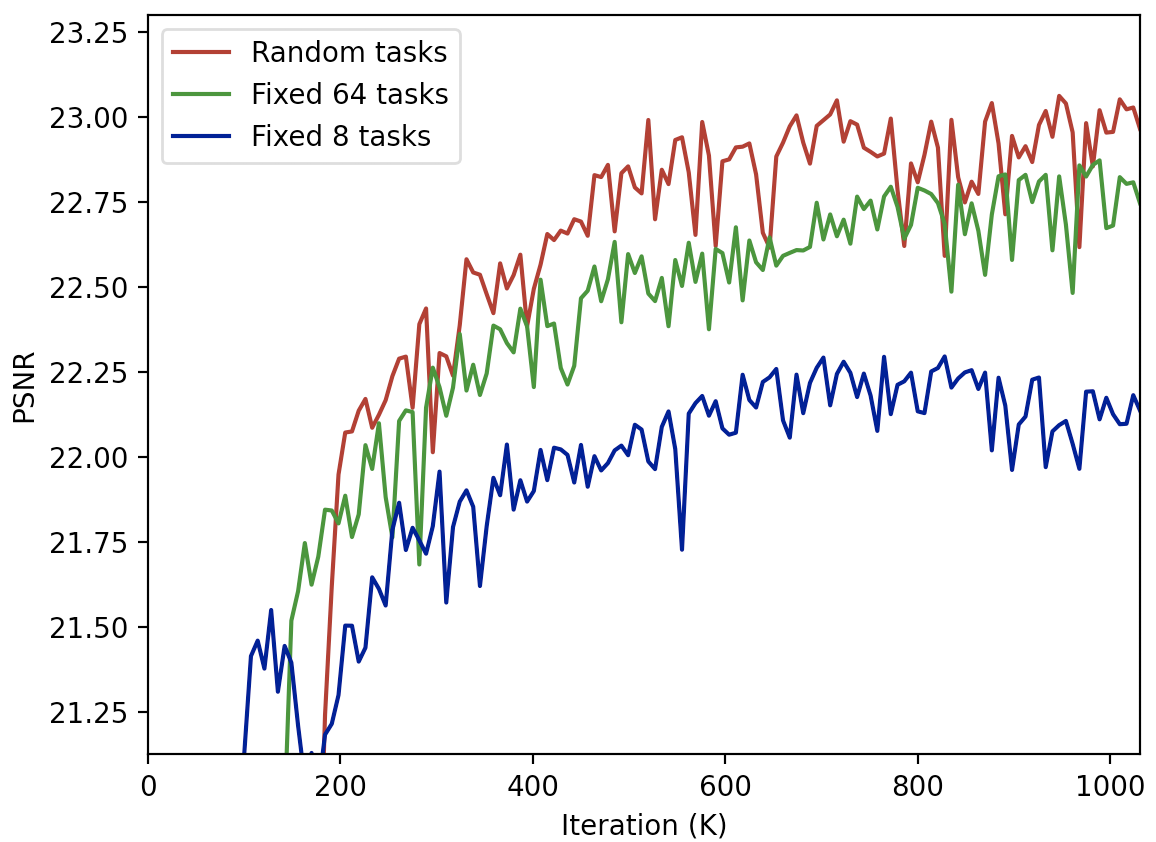}
  \centering
  \caption{PSNR curves comparison of \textit{Severe} degradation with fixed 8 tasks, 64 tasks and random infinite tasks.}
  \label{RandomTask}
\end{figure}

\subsubsection{The Importance of Randomly Selected Tasks from A Infinite Degradation Pool }
%Why we use the random and infinite meta-tasks for each iteration, but not fix them with the specific task number at the beginning of training?
Why we use the random selected tasks from the infinite degradation pool for each iteration, but not fix them from a specific degradation pool at the beginning of training? We conduct experiments by using fixed 8 tasks, 64 tasks and random infinite tasks. The curves of PSNR results with \textit{Severe} degradation are shown in Fig~\ref{RandomTask}. As introduced in the experimental setting, the degradation distributions $p(\mathcal{T})$ of meta-training data is randomly sampled. For each training iteration with scale factor $s$, we sample $k$ tasks, where the blur kernel widths $\sigma_{G}$ are in the range of [0.2, $s$] and the noise levels $\sigma_{N}$ are in the range of [0, 75]. Some previous meta-network frameworks~\cite{finn2017model,zhang2020adaptive} fix the tasks at the beginning of training and feed all the tasks for one iteration. However, our framework cannot use the fix tasks because of the following two reasons: (1) we cannot define a certain number of tasks because the degradations of LR images are various and infinite; and (2) the fixed tasks will cause overfitting, which limits the generalization of CMDSR. As shown in Fig~\ref{RandomTask}, we can observe that both models trained with fixed tasks get worse performance with distribution shifts. Moreover, when using 8 tasks, the curve of training loss even shows performance drop. The model trained with random tasks achieves the best performance at shifted degradation, which indicates that the setting of random tasks is significant for our framework.

\subsubsection{The Number of Task Size, Support Size and Patch Size} To evaluate the influence of task size $k$, support size $n$ and patch size $h\times w$, we conduct experiments by changing the number of task size ($k=2,4,8$), support size ($n=1,10,15,20,25,30$) and patch size ($h\times w=8\times 8,32\times 32,48\times 48$).

First, we fix the support size and increase the task size. The number of task size defines the number of tasks which are used for each training step. When we use a larger task size, more tasks will be fed into CMDSR for each iteration. As Table~\ref{task size} shows, the CMDSR benefits from larger training task size.
\begin{table}[htbp]
  % \vspace{-1.1cm}
  % \setlength{\belowcaptionskip}{-0.7cm}
% \begin{center}
\centering
\caption{Average $\times$4 PSNR on Set5 with \textit{Middle} degradation, where CMDSR is trained with different task sizes.}
\label{task size}
\centering
\begin{tabular}{cccc}
\hline
Model&Task Size&Support Size&PSNR\\
\hline
\hline
\multirow{3}{*}{CMDSR}& 2& 20& 26.96\\
& 4& 20& 27.03\\
& 8& 20& \textcolor{red}{27.10}\\
\hline
\end{tabular}
% \end{center}
\end{table}

\begin{figure}[h]
  \includegraphics[width=7cm]{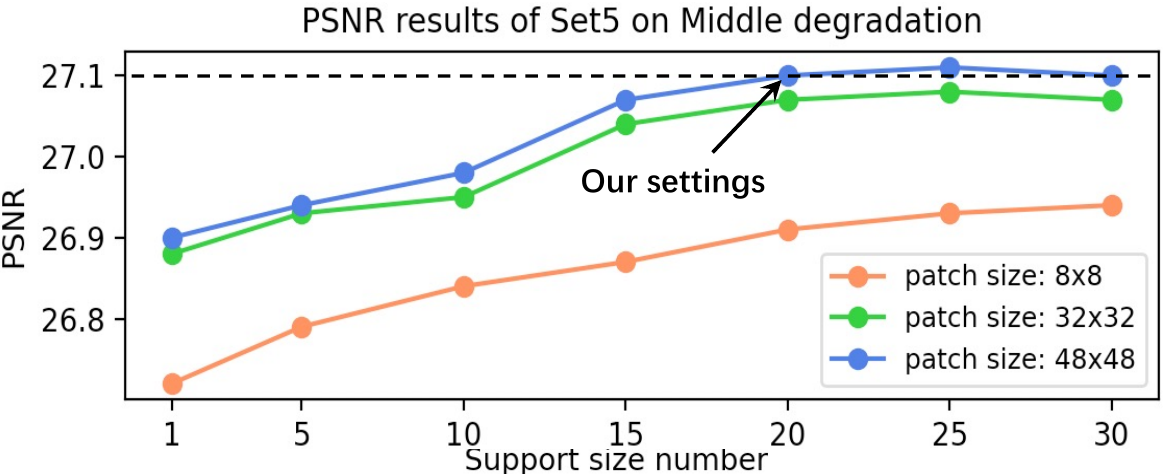}
  \centering
  \caption{PSNRs on \textit{Middle} Set5 using different support sizes and patch sizes.}
  \label{support_patch_size}
\end{figure}

Then, we fix the task size and change the support size and patch size. The number of support size defines the number of LR images of the support set for each task. As Fig.~\ref{support_patch_size} shows, the CMDSR benefits from larger support size and larger patch size. According to the network structure of ConditionNet, the number of support size also defines the input channel of ConditionNet. Larger support size and larger patch size is beneficial for ConditionNet to extract the degradation information more accurately from LR patch samples at task level. And our settings (support set size=20, patch size=48) are the most appropriate because further larger sizes bring only marginal improvements.

\subsubsection{Combination of Loss Functions} We compare the results of CMDSR, where ConditionNet is separately trained with task contrastive loss $L_{con}$, reconstruction loss $L_{res}$ and combined loss. The BaseNet is still optimized by the reconstruction loss $L_{res}$ for SR. As shown in Table~\ref{Combination of Loss Functions.}, if ConditionNet is trained with $L_{res}$, CMDSR gets collapsed to produce even worse results than the single BaseNet. Only when using the unsupervised $L_{con}$, the result is acceptable, but using the combined loss achieves the best results; This is because it makes a balance between task-level feature extraction and the generalization of SISR.

\begin{table}[t]
\centering
\caption{Average $\times$4 PSNR on Set5 with \textit{Middle} degradation, where ConditionNet is separately trained with three losses.}
\label{Combination of Loss Functions.}
\centering
\begin{tabular}{cccc}
\hline
Model&\makecell[c]{Loss of\\ BaseNet}&\makecell[c]{Loss of\\ ConditionNet}&PSNR\\
\hline
\hline
\multirow{3}{*}{\makecell[c]{ConditionNet + BaseNet\\ (CMDSR)}}&$L_{res}$&$L_{res}$&23.36\\
&$L_{res}$&$L_{con}$&26.84\\
&$L_{res}$&$L_{con}+\lambda *L_{res}$&\textcolor{red}{27.10}\\
\hline
\end{tabular}
\end{table}

\subsubsection{The Effect of Contrastive Loss and Reconstruction Loss Ratio} How to balance the effect of Contrastive Loss and Reconstruction Loss to better optimize the ConditionNet? We conduct experiments by changing the loss coefficient ($\lambda=1, 0.1, 0.01$). The PSNR results with \textit{Middle} degradation are shown in Table~\ref{Loss Ratio}. It is observed that the large ratio of Reconstruction loss ($\lambda=1$) is harmful for the task-level feature extraction of ConditionNet. When we decrease the ratio ($\lambda=0.01$), the performance of CMDSR is slightly decreased. Therefore, we set the loss coefficient $\lambda$ as 0.1.
\begin{table}[htbp]
  % \vspace{-1.1cm}
  % \setlength{\belowcaptionskip}{-0.7cm}
% \begin{center}
\centering
\caption{Average $\times$4 PSNR on Set5 with \textit{Middle} degradation, where ConditionNet is separately trained with different loss coefficients.}
\label{Loss Ratio}
\centering
\begin{tabular}{cccc}
\hline
Model&\makecell[c]{Combined Loss of\\ ConditionNet}&coefficient&PSNR\\
\hline
\hline
\multirow{3}{*}{CMDSR}& \multirow{3}{*}{$L_{con}+\lambda *L_{res}$}& $\lambda = 1$& 25.27\\
& & $\lambda = 0.1$& \textcolor{red}{27.10}\\
& & $\lambda = 0.01$& 27.03\\
\hline
\end{tabular}
% \end{center}
\end{table}

\subsubsection{The Importance of Alternating Training Strategy}
As we have explained before, the ConditionNet and BaseNet serve different purposes with different learning rates and loss functions. Here, we have conducted the experiment to show the importance of  alternating training strategy with different steps $t_0$.  Specifically, we alternately optimize the BaseNet and ConditionNet with different steps ($t_0=0, 1,10, 20)$. The results are presented in Table~\ref{step}. It can be seen that the alternative training strategy is quite important, since non-alternative training $t_0=0$ produces the worst performance. We speculate that this is caused by the inconsistent convergence speed of the  ConditionNet and Basenet. To make a balance between the conditional feature extraction and SR restoration, we experimentally set the alternating training step $t_0=10$ to train ConditionNet and BaseNet alternatively.

\begin{table}[htbp]
\centering
\caption{Average $\times$4 PSNR on Set5 with \textit{Middle} degradation using different alternating training step $t_0$.}
\label{step}
\centering
\begin{tabular}{ccc}
\hline
Model&Alternating training step $t_0$&PSNR\\
\hline
\hline
\multirow{4}{*}{CMDSR}&0&26.86\\
&1&27.05\\
&10&\color{red}{27.10}\\
&20&27.01\\
\hline
\end{tabular}
\end{table}

\subsubsection{Can CMDSR Use Other Modulation?}
Since our model aims to be a general framework, we use depth-wise scaling to adapt convolutions without changing the CNN structures or involving more network parameters. Hence, we applied the depth-wise scaling as our modulation. Here, we extend to use the novel  Spatial Feature Transform (SFT) of IKC~\cite{gu2019blind} as the modulation. As shown in Table~\ref{Modulation}, CMDSR can further benefit from SFT which concatenates the degradation and image features to apply an affine transformation in each middle layer. However, after using the SFT modulations, the parameters of CMDSR get greatly increased. With sufficient calculation resources, we believe more novel modulations can be applied to our framework in the future.

\begin{table}[htbp]
\centering
\caption{Average $\times$4 PSNR on Set5 with \textit{Middle} degradation using different modulations.}
\label{Modulation}
\centering
\begin{tabular}{cccc}
\hline
Model&Modulation&PSNR&Params.\\
\hline
\hline
\multirow{2}{*}{CMDSR}&depth-wise scaling&27.10& 1.48M\\
& SFT~\cite{gu2019blind}& 27.53 $\uparrow$& 5.92M\\
\hline
\end{tabular}
\end{table}

\subsubsection{Can CMDSR Extend to Other SISR Structure?}
As mentioned before, since our model uses depth-wise scaling to adapt convolutions without changing CNN structures, it has no strict restrictions on the BaseNet structure. Therefore, we attempt to extend our framework to other SISR structure. Specifically, we replace SRResNet-10 with other three SISR models, VDSR\cite{kim2016accurate}, EDSR\cite{lim2017enhanced} and IDN~\cite{hui2018fast}. Moreover, we also use the same training data described in Section~\ref{Experimental Setups} to train these models without ConditionNet. As shown in Table~\ref{other structures}, all joint models obtain significant improvement, and EDSR~\cite{zhang2018image} achieves the best results with the largest number of parameters. We believe that our framework can be extended to more complicated structures in the future.

\begin{table}[htbp]
\centering
\caption{Average $\times$4 PSNR on Set5 with \textit{Middle} degradation using other structures for BaseNet.}
\label{other structures}
\centering
\begin{tabular}{ccc}
\hline
Model&\makecell[c]{BaseNet\\ Parameters}&PSNR\\
\hline
\hline
VDSR~\cite{kim2016accurate} w/o ConditionNet & \multirow{2}{*}{0.67M}& 26.49\\
VDSR~\cite{kim2016accurate} w/ ConditionNet & & 26.97$\uparrow$\\
\hline
IDN~\cite{hui2018fast} w/o ConditionNet & \multirow{2}{*}{0.80M}& 26.53\\
IDN~\cite{hui2018fast} w/ ConditionNet & & 27.03$\uparrow$\\
\hline
EDSR~\cite{lim2017enhanced} w/o ConditionNet & \multirow{2}{*}{43M}& 26.81\\
EDSR~\cite{lim2017enhanced} w/ ConditionNet & & 27.51$\uparrow$\\
\hline
\end{tabular}
\end{table}

\section{Conclusion}
In this paper, we investigate the blind SISR problem with multiple degradations. Inspired by meta-learning, we design the first blind SR hyper-network that learns how to adapt to changes in input distribution. Specifically, we use a ConditionNet to extract the task-level features with batches of LR patches and use BaseNet to rapidly adapt its parameters according to the conditional features. Extensive experiments show that our framework can handle distribution shift by only performing one-step adaptation. For complicated and real scenes, our blind model even outperforms several non-blind models. Besides, our framework is flexible and efficient so that we can extend the structure of BaseNet to other SISR models.  %In summary, the proposed framework
In future work, we will extend our general framework to more CNN models and more low-level vision tasks.

\section{Acknowledgement}
This research is supported by the Sea-NExT Joint Lab and ByteDance AI Lab.

\ifCLASSOPTIONcaptionsoff
  \newpage
\fi

\bibliographystyle{IEEEtran}
\bibliography{egbib}
\end{document}